\pdfoutput=1
\documentclass[11pt,a4paper]{article}
\usepackage[hyperref]{emnlp-ijcnlp-2019}
\usepackage{times}
\usepackage{latexsym}

\usepackage{adjustbox}
\usepackage{verbatim}
\usepackage{graphicx}
\usepackage{amstext}
\usepackage{amsmath}
\DeclareGraphicsExtensions{.pdf,.png,.jpg}

\usepackage{url}
\usepackage{xcolor}
\usepackage{cancel}

\aclfinalcopy 

\title{Summary Level Training of Sentence Rewriting \\ for Abstractive Summarization}

\author{Sanghwan Bae, Taeuk Kim, Jihoon Kim {\normalfont and} Sang-goo Lee \\
  Department of Computer Science and Engineering \\
  Seoul National University, Seoul, Korea \\
  {\tt \{sanghwan,taeuk,kjh255,sglee\}@europa.snu.ac.kr} \\}

\date{}

\begin{document}
\maketitle
\begin{abstract}
  As an attempt to combine extractive and abstractive summarization, \emph{Sentence Rewriting} models adopt the strategy of extracting salient sentences from a document first and then paraphrasing the selected ones to generate a summary.
  However, the existing models in this framework mostly rely on sentence-level rewards or suboptimal labels, 
  causing a mismatch between a training objective and evaluation metric.
  In this paper, we present a novel training signal that directly maximizes summary-level ROUGE scores through reinforcement learning. In addition, we incorporate BERT into our model, making good use of its ability on natural language understanding. 
  In extensive experiments, we show that a combination of our proposed model and training procedure obtains new state-of-the-art performance on both CNN/Daily Mail and New York Times datasets.
  We also demonstrate that it generalizes better on DUC-2002 test set.
\end{abstract}

\section{Introduction}

The task of automatic text summarization aims to
compress a textual document to a shorter highlight
while keeping salient information of the original text.
In general, there are two ways to do text summarization:
\emph{Extractive} and \emph{Abstractive} \cite{mani2001automatic}.
\emph{Extractive} approaches generate summaries by
selecting salient sentences or phrases from 
a source text, while \emph{abstractive} approaches
involve a process of paraphrasing or generating sentences to write a summary.

Recent work \cite{liu2019fine, zhang-etal-2019-hibert} demonstrates that it is highly beneficial for extractive summarization models to incorporate pre-trained language models (LMs) such as BERT \cite{devlin-etal-2019-bert} into their architectures. 
However, the performance improvement from the pre-trained LMs is known to be relatively small in case of abstractive summarization \cite{zhang2019pretraining, hoang2019efficient}.
This discrepancy may be due to the difference between extractive and abstractive approaches in ways of dealing with the task---the former \textit{classifies} whether each sentence to be included in a summary, while the latter \textit{generates} a whole summary from scratch. 
In other words, as most of the pre-trained LMs are designed to be of help to the tasks which can be categorized as classification including extractive summarization, they are not guaranteed to be advantageous to abstractive summarization models that should be capable of generating language \cite{wang-cho-2019-bert, zhang2019bertscore}.


On the other hand, recent 
studies for abstractive summarization \cite{chen-bansal-2018-fast, hsu-etal-2018-unified, gehrmann-etal-2018-bottom} have attempted to exploit extractive models.
Among these, a notable one is \citet{chen-bansal-2018-fast}, in which a sophisticated model called \emph{Reinforce-Selected Sentence Rewriting} is proposed.
The model consists of both an extractor and abstractor, where the extractor picks out salient sentences first from a source article, and then the abstractor rewrites and compresses the extracted sentences into a complete summary. 
It is further fine-tuned by training the extractor with the rewards derived from sentence-level ROUGE scores of the summary generated from the abstractor.


In this paper, we improve the model of \citet{chen-bansal-2018-fast}, addressing two primary issues.
Firstly, we argue there is a bottleneck in the existing extractor on the basis of the observation that its performance as an independent summarization model (i.e., without the abstractor) is no better than solid baselines such as selecting the first 3 sentences.
To resolve the problem, we present a novel neural extractor exploiting the pre-trained LMs (BERT in this work) which are expected to perform better according to the recent studies \cite{liu2019fine, zhang-etal-2019-hibert}.
Since the extractor is a sort of sentence classifier, we expect that it can make good use of the ability of pre-trained LMs which is proven to be effective in classification.


Secondly, the other point is that there is a mismatch between the training objective and evaluation metric; the previous work utilizes the \emph{sentence-level} ROUGE scores as a reinforcement learning objective, while the final performance of a summarization model is evaluated by the \emph{summary-level} ROUGE scores.
Moreover, as \citet{narayan-etal-2018-ranking} pointed out, sentences with the highest individual ROUGE scores do not necessarily lead to an optimal summary,
since they may contain overlapping contents, causing verbose and redundant summaries. 
Therefore, we propose to directly use the summary-level ROUGE scores as an objective instead of the sentence-level scores.
A potential problem arising from this apprsoach is the sparsity of training signals, because the summary-level ROUGE scores are calculated only once for each training episode.
To alleviate this problem, we use \emph{reward shaping} \cite{Ng:1999:PIU:645528.657613} to give an intermediate signal for each action, preserving the optimal policy.




We empirically demonstrate the superiority of our approach by achieving new state-of-the-art abstractive summarization results on CNN/Daily Mail and New York Times datasets \cite{NIPS2015_5945, durrett-etal-2016-learning}. It is worth noting that our approach shows large improvements especially on ROUGE-L score which is considered a means of assessing fluency \cite{narayan-etal-2018-ranking}.
In addition, our model performs much better than previous work when testing on DUC-2002 dataset, showing better generalization and robustness of our model.


Our contributions in this work are three-fold: 
a novel successful application of pre-trained transformers for abstractive summarization;
suggesting a training method to globally optimize sentence selection;
achieving the state-of-the-art results on the benchmark datasets, CNN/Daily Mail and New York Times.


\begin{figure*}[t]
\centering
\begin{center}
    \includegraphics[width=0.95\textwidth]{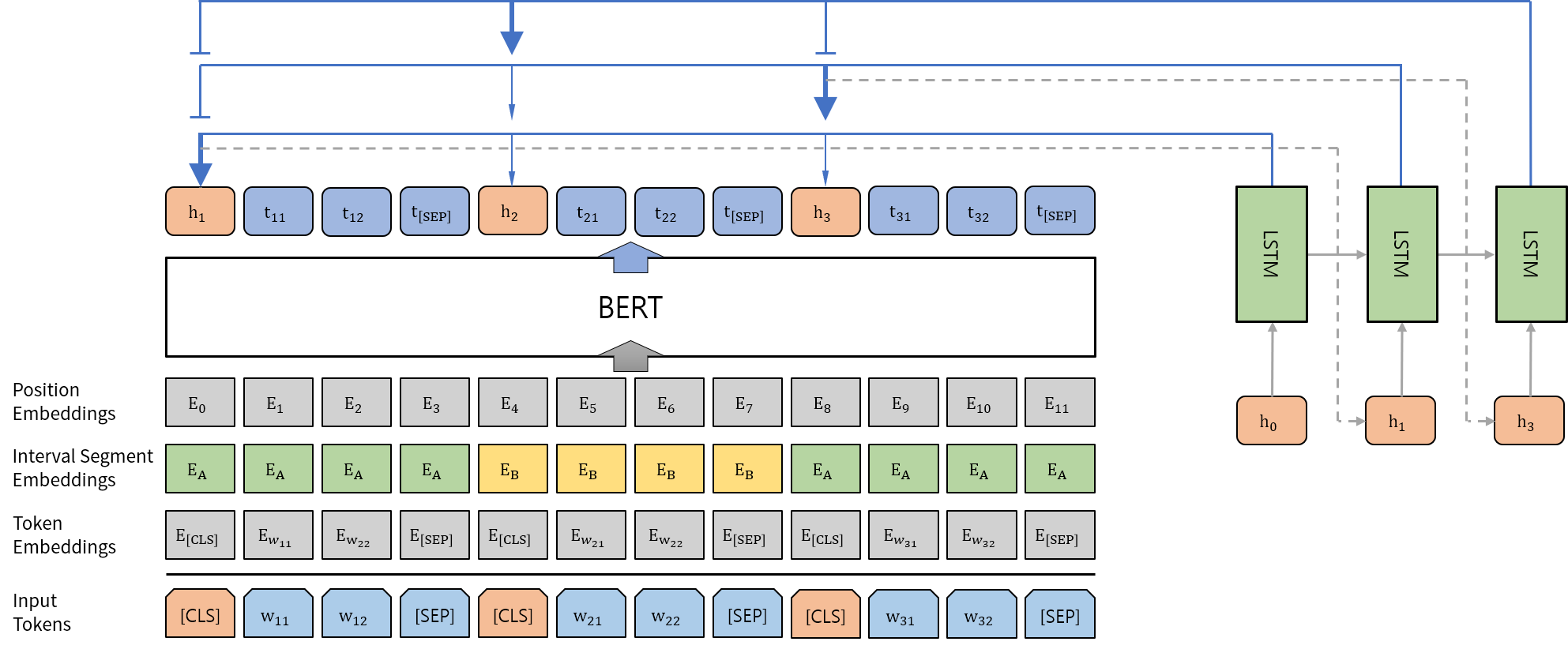}
    \caption{The overview architecture of the extractor netwrok}
    \label{fig:extractor_full}
\end{center}
\end{figure*}

\section{Background}

\subsection{Sentence Rewriting}

In this paper, we focus on single-document multi-sentence summarization and propose
a neural abstractive model based on the \emph{Sentence Rewriting} framework 
\cite{chen-bansal-2018-fast, xu2019neural} which consists of two parts:
a neural network for the \emph{extractor} and another network for the \emph{abstractor}.
The extractor network is designed to extract salient sentences from a source article.
The abstractor network rewrites the extracted sentences
into a short summary.

\subsection{Learning Sentence Selection}
\label{sec:optimality}

The most common way to train extractor to select
informative sentences is building extractive oracles
as gold targets, and training with cross-entropy (CE) loss.
An oracle consists of a set of sentences with the highest possible
ROUGE scores. Building oracles is finding an optimal combination of sentences,
where there are $2^n$ possible combinations for each example.
Because of this, the exact optimization for ROUGE scores is intractable. 
Therefore, alternative methods identify the set of sentences
with greedy search \cite{nallapati2017summarunner},
sentence-level search \cite{hsu-etal-2018-unified, shi2019deepchannel}
or collective search using the limited number of sentences \cite{xu2019neural},
which construct suboptimal oracles.
Even if all the optimal oracles are found,
training with CE loss using these labels will cause underfitting
as it will only maximize probabilities for sentences
in label sets and ignore all other sentences.

Alternatively, reinforcement learning (RL) can give room for exploration in the search space.
\citet{chen-bansal-2018-fast}, our baseline work,
proposed to apply policy gradient methods to train an extractor.
This approach makes an end-to-end trainable stochastic
computation graph, encouraging the model to select sentences
with high ROUGE scores. However, they define a reward for an action
(sentence selection) as a sentence-level ROUGE score between
the chosen sentence and a sentence in the ground truth summary
for that time step. This leads the extractor agent to a suboptimal policy;
the set of sentences matching individually with each sentence in a ground truth summary
isn't necessarily optimal in terms of summary-level ROUGE score.

\citet{narayan-etal-2018-ranking} proposed policy gradient
with rewards from summary-level ROUGE.
They defined an action as sampling a summary from candidate summaries
that contain the limited number of plausible sentences. After training,
a sentence is ranked high for selection
if it often occurs in high scoring summaries.
However, their approach still has a risk of ranking redundant sentences high;
if two highly overlapped sentences have salient information,
they would be ranked high together, increasing the probability of being sampled
in one summary.

To tackle this problem, we propose a training method
using reinforcement learning which globally optimizes
summary-level ROUGE score and gives intermediate rewards
to ease the learning.


\subsection{Pre-trained Transformers}

Transferring representations from pre-trained transformer language models
has been highly successful in the domain of natural language understanding tasks 
\cite{radford2018improving, devlin-etal-2019-bert, radford2019language, yang2019xlnet}.
These methods first pre-train highly stacked transformer blocks \cite{vaswani2017attention}
on a huge unlabeled corpus, and then fine-tune the models or representations
on downstream tasks.

%
%





\section{Model}

Our model consists of two neural network modules, i.e. an extractor and abstractor.
The extractor encodes a source document and chooses
sentences from the document, and then the abstractor
paraphrases the summary candidates.
Formally, a single document consists of $n$ sentences $D=\{s_1,s_2,\cdots,s_n\}$.
We denote $i$-th sentence as $s_i=\{w_{i1},w_{i2},\cdots,w_{im}\}$ where $w_{ij}$ is
the $j$-th word in $s_i$.
The extractor learns to pick out a subset of $D$ denoted as
$\hat{D}=\{\hat{s}_1,\hat{s}_2,\cdots,\hat{s}_k|\hat{s}_i\in D\}$
where $k$ sentences are selected. The abstractor rewrites each of the selected sentences
to form a summary $S=\{f(\hat{s}_1),f(\hat{s}_2),\cdots,f(\hat{s}_k)\}$,
where $f$ is an abstracting function.
And a gold summary consists of $l$ sentences $A=\{a_1,a_2,\cdots,a_l\}$.

\subsection{Extractor Network}

The extractor is based on the encoder-decoder framework.
We adapt BERT for the encoder to exploit contextualized
representations from pre-trained transformers. BERT as the
encoder maps the input sequence $D$ to sentence representation
vectors $H=\{h_1,h_2,\cdots,h_n\}$,
where $h_i$ is for the $i$-th sentence in the document.
Then, the decoder utilizes $H$ to extract $\hat{D}$ from $D$.

\subsubsection{Leveraging Pre-trained Transformers}

Although we require the encoder
to output the representation for each sentence,
the output vectors from BERT are grounded to tokens
instead of sentences. Therefore, we modify the input
sequence and embeddings of BERT as \citet{liu2019fine} did.

In the original BERT's configure, a [CLS] token is used
to get features from one sentence or a pair of sentences.
Since we need a symbol for each sentence representation,
we insert the [CLS] token before each sentence. And we
add a [SEP] token at the end of each sentence, which is used to
differentiate multiple sentences. As a result, the vector for the $i$-th 
[CLS] symbol from the top BERT layer corresponds to the $i$-th sentence representation $h_i$.

In addition, we add interval segment embeddings as input for BERT
to distinguish multiple
sentences within a document. For $s_i$ we assign
a segment embedding $E_A$ or $E_B$ conditioned on $i$
is odd or even. For example, for a consecutive sequence of sentences $s_1, s_2, s_3, s_4, s_5$, 
we assign $E_A, E_B, E_A, E_B, E_A$ in order.
All the words in each sentence are assigned to the same
segment embedding, \emph{i.e.} segment embeddings for
$w_{11}, w_{12},\cdots,w_{1m}$ is $E_A,E_A,\cdots,E_A$.
An illustration for this procedure is shown in Figure \ref{fig:extractor_full}.

\subsubsection{Sentence Selection}

We use LSTM Pointer Network \cite{vinyals2015pointer}
as the decoder to select the extracted sentences
based on the above sentence representations. The decoder
extracts sentences recurrently, producing a distribution
over all of the remaining sentence representations
excluding those already selected.
Since we use the sequential model which selects
one sentence at a time step, our decoder
can consider the previously selected sentences.
This property is needed to avoid selecting sentences
that have overlapping information
with the sentences extracted already.

As the decoder structure is almost the same with the previous work,
we convey the equations of \citet{chen-bansal-2018-fast}
to avoid confusion, with minor modifications to agree with our notations.
Formally, the extraction probability is calculated as:
\begin{gather}
\label{eq:extractor_score}
u_{t,i} = v_m^\top\tanh(W_ee_t + W_hh_i) \\
P(\hat{s}_t|D,\hat{s}_1,\cdots,\hat{s}_{t-1})=\mathrm{softmax}(u_t)
\end{gather}
where $e_t$ is the output of the glimpse operation:
\begin{gather}
\label{eq:glimpse_operation}
c_{t,i}=v_g^\top\tanh(W_{g1}h_i+W_{g2}z_t) \\
\alpha_t=\mathrm{softmax}(c_t) \\
e_t=\sum_i\alpha_tW_{g1}h_i
\end{gather}
In Equation \ref{eq:glimpse_operation}, $z_t$ is the hidden state
of the LSTM decoder at time $t$ (shown in green in Figure \ref{fig:extractor_full}).
All the $W$ and $v$ are trainable parameters.

\subsection{Abstractor Network}
The abstractor network approximates $f$, which compresses
and paraphrases an extracted document sentence to a concise
summary sentence. We use the standard attention based
sequence-to-sequence (seq2seq) model \cite{BahdanauCB14, luong-etal-2015-effective}
with the copying mechanism \cite{see-etal-2017-get}
for handling out-of-vocabulary (OOV) words.
Our abstractor is practically identical to the one
proposed in \citet{chen-bansal-2018-fast}.

\section{Training} \label{Training}

In our model, an extractor selects a series of sentences, and then an abstractor paraphrases them. As they work in different ways,
we need different training strategies suitable for each of them. Training the abstractor
is relatively obvious; maximizing log-likelihood for the next word
given the previous ground truth words. However, there are several
issues for extractor training. First, the extractor should consider
the abstractor's rewriting process when it selects sentences.
This causes a \emph{weak supervision} problem \cite{jehl-etal-2019-neural},
since the extractor gets training signals indirectly
after paraphrasing processes are finished.
In addition, thus this procedure contains sampling or maximum selection,
the extractor performs a non-differentiable extraction.
Lastly, although our goal is maximizing ROUGE scores,
neural models cannot be trained directly by maximum likelihood estimation from them.

To address those issues above, we apply standard policy gradient methods,
and we propose a novel training procedure for extractor
which guides to the optimal policy in terms of the summary-level ROUGE. As usual in RL for sequence prediction,
we pre-train submodules and apply RL to fine-tune the extractor.

\subsection{Training Submodules}
\label{sec:submodule}

\paragraph{Extractor Pre-training}

Starting from a poor random policy makes it difficult to
train the extractor agent to converge towards the optimal policy.
Thus, we pre-train the network using cross entropy (CE) loss
like previous work \cite{DBLP:conf/iclr/BahdanauBXGLPCB17, chen-bansal-2018-fast}.
However, there is no gold label for extractive summarization
in most of the summarization datasets.
Hence, we employ a greedy approach \cite{nallapati2017summarunner}
to make the extractive oracles,
where we add one sentence at a time incrementally to the summary,
such that the ROUGE score of the current set of selected sentences
is maximized for the entire ground truth summary.
This doesn't guarantee optimal, but it is enough
to teach the network to select plausible sentences.
Formally, the network is trained to minimize
the cross-entropy loss as follows:
\begin{equation}
\label{eq:}
L_{\text{ext}}=-\frac{1}{T}\sum_{t=1}^{T}\log P(s^*_t|D,s^*_1,\cdots,s^*_{t-1})
\end{equation}
where $s^*_t$ is the $t$-th generated oracle sentence.

\paragraph{Abstractor Training}

For the abstractor training, we should create
training pairs for input and target sentences.
As the abstractor paraphrases on sentence-level,
we take a sentence-level search for each ground-truth
summary sentence. We find the most similar document
sentence $s'_t$ by:
\begin{equation}
\label{eq:}
s'_t = \text{argmax}_{s_i}(\text{ROUGE-L}^{\text{sent}}_{F_1}(s_i,a_t))
\end{equation}
And then the abstractor is trained as a usual sequence-to-sequence
model to minimize  the cross-entropy loss:
\begin{equation}
\label{eq:}
L_{\text{abs}}=-\frac{1}{m}\sum_{j=1}^m\log P(w^a_j|w^a_1,\cdots,w^a_{j-1},\Phi)
\end{equation}
where $w^a_j$ is the $j$-th word of the target sentence $a_t$,
and $\Phi$ is the encoded representation for $s'_t$.

\subsection{Guiding to the Optimal Policy}

To optimize ROUGE metric directly, we assume the extractor
as an agent in reinforcement learning paradigm \cite{sutton1998introduction}.
We view the extractor has a stochastic \emph{policy} that generates
\emph{actions} (sentence selection) and receives the score of final evaluation metric
(summary-level ROUGE in our case) as the \emph{return}
\begin{equation}
\label{eq:}
R(S)=\text{ROUGE-L}^{\text{summ}}_{F_1}(S, A).
\end{equation}

While we are ultimately interested in the maximization of the
score of a complete summary, simply awarding this score at the
last step provides a very sparse training signal.
For this reason we define intermediate rewards using
\emph{reward shaping} \cite{Ng:1999:PIU:645528.657613},
which is inspired by \citet{DBLP:conf/iclr/BahdanauBXGLPCB17}'s attempt for sequence prediction.
Namely, we compute summary-level score values for all intermediate
summaries:
\begin{equation}
\label{eq:}
(R(\{\hat{s}_1\}),R(\{\hat{s}_1,\hat{s}_2\}),\cdots,R(\{\hat{s}_1,\hat{s}_2,\cdots,\hat{s}_k\}))
\end{equation}
The reward for each step $r_t$ is the difference between the consecutive pairs of scores:
\begin{equation}
\label{eq:}
r_t = R(\{\hat{s}_1,\hat{s}_2,\cdots,\hat{s}_t\}) - R(\{\hat{s}_1,\hat{s}_2,\cdots,\hat{s}_{t-1}\})
\end{equation}
This measures an amount of increase or decrease in the
summary-level score from selecting $\hat{s}_t$.
Using the shaped reward $r_t$ instead of awarding the whole score
$R$ at the last step does not change the optimal policy \cite{Ng:1999:PIU:645528.657613}.
We define a discounted future reward for each step as
$R_t=\sum_{t=1}^{k}\gamma^tr_{t+1}$, where $\gamma$ is a discount factor.

Additionally, we add `stop' action to the action space, by concatenating
trainable parameters $h_{\text{stop}}$ (the same dimension as $h_i$)
to $H$. The agent treats it as another candidate to extract. When it selects
`stop', an extracting episode ends and the final return is given.
This encourages the model to extract additional sentences only when they
are expected to increase the final return.

Following \citet{chen-bansal-2018-fast}, we use the
\emph{Advantage Actor Critic} \cite{mnih2016asynchronous} method to train.
We add a critic network to estimate a value function
$V_t(D,\hat{s}_1,\cdots,\hat{s}_{t-1})$, which
then is used to compute advantage of each action
(we will omit the current state $(D,\hat{s}_1,\cdots,\hat{s}_{t-1})$ to simplify):
\begin{equation}
\label{eq:advantage}
A_t(s_i)=Q_t(s_i) - V_t.
\end{equation}
where $Q_t(s_i)$ is the expected future reward for selecting $s_i$ at the current step $t$.
We maximize this advantage with the policy gradient with the Monte-Carlo sample
($A_t(s_i) \approx R_t - V_t$):
\begin{gather}
\label{eq:policy_gradient}
\begin{adjustbox}{max width=\columnwidth} {
$\nabla_{\theta_\pi}L_{\pi}\approx
\frac{1}{k}\sum_{t=1}^{k}\nabla_{\theta_\pi}\log{P(s_i|D,\hat{s}_1,\cdots,\hat{s}_{t-1})}
A_t(s_i)$}
\end{adjustbox}
\end{gather}
where $\theta_\pi$ is the trainable parameters of the actor network (original extractor).
And the critic is trained to minimize the square loss:
\begin{equation}
\label{eq:critic_loss}
\nabla_{\theta_\psi}L_{\psi}= \nabla_{\theta_\psi}(V_t - R_t)^2
\end{equation}
where $\theta_\psi$ is the trainable parameters of the critic network.

\begin{table*}[t!]
\centering
\begin{adjustbox}{max width=\textwidth}
\begin{tabular}{l|ccc|c}
    \hline
    \bf Models & \bf ROUGE-1 & \bf ROUGE-2 & \bf ROUGE-L & \bf R-AVG \\
    \hline
    \multicolumn{5}{l}{Extractive} \\
    \hline
    lead-3 \cite{see-etal-2017-get} & 40.34 & 17.70 & 36.57 & 31.54 \\
    REFRESH \cite{narayan-etal-2018-ranking} & 40.00 & 18.20 & 36.60 & 31.60 \\
    JECS \cite{xu2019neural} & 41.70 & 18.50 & 37.90 & 32.70 \\
    HiBERT \cite{zhang-etal-2019-hibert} & 42.37 & 19.95 & 38.83 & 33.71 \\
    BERTSUM \cite{liu2019fine} & \bf 43.25 & \bf 20.24 & 39.63 & \bf 34.37 \\
    BERT-ext (ours) & 42.29 & 19.38 & 38.63 & 33.43 \\
    BERT-ext + RL (ours) & 42.76 & 19.87 & 39.11 & 33.91 \\
    \hline
    \multicolumn{5}{l}{Abstractive} \\
    \hline
    Pointer Generator \cite{see-etal-2017-get} & 39.53 & 17.28 & 36.38 & 31.06 \\
    Inconsistency Loss \cite{hsu-etal-2018-unified} & 40.68 & 17.97 & 37.13 & 31.93 \\
    Sentence Rewrite (w/o rerank) \cite{chen-bansal-2018-fast} & 40.04 & 17.61 & 37.59 & 31.74 \\
    Sentence Rewrite \cite{chen-bansal-2018-fast} & 40.88 & 17.80 & 38.54 & 32.41 \\
    Bottom-Up \cite{gehrmann-etal-2018-bottom} & 41.22 & 18.68 & 38.34 & 32.75 \\
    Transformer-LM \cite{hoang2019efficient} & 38.67 & 17.47 & 35.79 & 30.64 \\
    Two-Stage BERT \cite{zhang2019pretraining} & 41.71 & \bf 19.49 & 38.79 & 33.33 \\
    BERT-ext + abs (ours) & 40.14 & 17.87 & 37.83 & 31.95 \\
    BERT-ext + abs + rerank (ours) & 40.71 & 17.92 & 38.51 & 32.38 \\
    BERT-ext + abs + RL (ours) & 41.00 & 18.81 & 38.51 & 32.77 \\
    BERT-ext + abs + RL + rerank (ours) & \bf 41.90 & 19.08 & \bf 39.64 & \bf 33.54 \\
    \hline
\end{tabular}
\end{adjustbox}
\caption{\label{table:cnndm} Performance on CNN/Daily Mail test set using the full length ROUGE $F_1$ score. R-AVG calculates average score of ROUGE-1, ROUGE-2 and ROUGE-L.}
\end{table*}

\section{Experimental Setup}

\subsection{Datasets}

We evaluate the proposed approach on the CNN/Daily Mail \cite{NIPS2015_5945} and
New York Times \cite{sandhaus2008nyt} dataset, which are both standard corpora
for multi-sentence abstractive summarization. Additionally, we test
generalization of our model on DUC-2002 test set.

CNN/Daily Mail dataset consists of more than 300K news articles and
each of them is paired with several highlights.
We used the standard splits of \citet{NIPS2015_5945}
for training, validation and testing (90,226/1,220/1,093 documents for CNN and
196,961/12,148/10,397 for Daily Mail). We did not anonymize entities.
We followed the preprocessing methods in \citet{see-etal-2017-get}
after splitting sentences by Stanford CoreNLP \cite{manning-EtAl:2014:P14-5}.

The New York Times dataset also consists of many news articles.
We followed the dataset splits of \citet{durrett-etal-2016-learning};
100,834 for training and 9,706 for test examples.
And we also followed the filtering procedure of them, removing
documents with summaries that are shorter than 50 words.
The final test set (NYT50) contains 3,452 examples out of the original 9,706.

The DUC-2002 dataset contains 567 document-summary pairs
for single-document summarization. As a single document can
have multiple summaries, we made one pair per summary.
We used this dataset as a test set for our model trained on
CNN/Daily Mail dataset to test generalization.

\subsection{Implementation Details}

Our extractor is built on $\text{BERT}_\text{BASE}$ with fine-tuning, smaller version
than $\text{BERT}_\text{LARGE}$ due to limitation of time and space.
We set LSTM hidden size as 256 for all of our models.
To initialize word embeddings for our abstractor,
we use word2vec \cite{mikolov2013distributed} of 128 dimensions
trained on the same corpus.
We optimize our model with Adam optimizer \cite{DBLP:journals/corr/KingmaB14} with
$\beta_1=0.9$ and $\beta_2=0.999$.
For extractor pre-training, we use learning rate schedule following \cite{vaswani2017attention}
with $warmup=10000$:
\begin{align*}
lr=2e^{-3}\cdot \min({steps^{-0.5}, steps\cdot warmup^{-1.5}}).
\end{align*}
And we set learning rate $1e^{-3}$ for abstractor and $4e^{-6}$ for RL training.
We apply gradient clipping using L2 norm with threshold $2.0$.
For RL training, we use $\gamma=0.95$ for the discount factor.
To ease learning $h_{\text{stop}}$,
we set the reward for the stop action to $\lambda\cdot\text{ROUGE-L}^{\text{summ}}_{F_1}(S, A)$,
where $\lambda$ is a stop coefficient set to $0.08$.
Our critic network shares the encoder with the actor (extractor)
and has the same architecture with it except the output layer,
estimating scalar for the state value. And the critic is initialized with
the parameters of the pre-trained extractor where it has the same architecture. 

\begin{table}
\begin{center}
\begin{adjustbox}{max width=0.95\columnwidth}
\begin{tabular}{l|ccc}
    \hline 
    \bf Models & \bf R-1 & \bf R-2 & \bf R-L \\ 
    \hline
    lead-3 \cite{see-etal-2017-get} & 40.34 & 17.70 & 36.57 \\
    rnn-ext \cite{chen-bansal-2018-fast} & 40.17 & 18.11 & 36.41 \\
    JECS-ext \cite{xu2019neural} & 40.70 & 18.00 & 36.80 \\ 
    BERT-ext (ours) & \bf 42.29 & \bf 19.38 & \bf 38.63 \\
    \hline
\end{tabular}
\end{adjustbox}
\end{center}
\caption{\label{table:extractor-baseline} Comparison of extractor networks.}
\end{table}

\subsection{Evaluation}

We evaluate the performance of our method using different variants
of ROUGE metric computed with respect to the gold summaries.
On the CNN/Daily Mail and DUC-2002 dataset, we use standard ROUGE-1,
ROUGE-2, and ROUGE-L \cite{lin-2004-rouge} on full length $F_1$ with stemming
as previous work did \cite{nallapati2017summarunner, see-etal-2017-get, chen-bansal-2018-fast}.
On NYT50 dataset,
following \citet{durrett-etal-2016-learning} and \citet{paulus2018a},
we used the limited length ROUGE recall metric, truncating the generated
summary to the length of the ground truth summary.

\section{Results}

\subsection{CNN/Daily Mail}

Table \ref{table:cnndm} shows the experimental results on
CNN/Daily Mail dataset, with extractive models in the top
block and abstractive models in the bottom block. For comparison,
we list the performance of many recent approaches with ours.

\paragraph{Extractive Summarization}
As \citet{see-etal-2017-get} showed, the first 3 sentences (lead-3)
in an article form a strong summarization baseline in CNN/Daily Mail dataset.
Therefore, the very first objective of extractive models is to outperform
the simple method which always returns 3 or 4 sentences at the top.
However, as Table \ref{table:extractor-baseline} shows,
ROUGE scores of lead baselines and extractors from previous work in
\emph{Sentence Rewrite} framework \cite{chen-bansal-2018-fast, xu2019neural}
are almost tie. We can easily conjecture that the limited performances of
their full model are due to their extractor networks.
Our extractor network with BERT (BERT-ext), as a single model,
outperforms those models with large margins.
Adding reinforcement learning (BERT-ext + RL) gives higher performance,
which is competitive with other
extractive approaches using pre-trained
Transformers (see Table \ref{table:cnndm}).
This shows the effectiveness of our learning method.

\begin{table}
\begin{center}
\begin{adjustbox}{max width=0.8\columnwidth}
\begin{tabular}{l|ccc}
    \hline
    & \bf R-1 & \bf R-2 & \bf R-L \\
    \hline
    Sentence-matching & 52.09 & 28.13 & 49.74 \\
    Greedy Search & 55.27 & 29.24 & 52.64 \\
    Combination Search & 55.51 & 29.33 & 52.89 \\
    \hline
\end{tabular}
\end{adjustbox}
\end{center}
\caption{\label{table:oracle} Comparison of different methods building upper bound
for full model.}
\end{table}

\paragraph{Abstractive Summarization}
Our abstractive approaches combine the extractor with the abstractor.
The combined model (BERT-ext + abs) without additional RL training
outperforms the Sentence Rewrite model
\cite{chen-bansal-2018-fast} without reranking, showing the
effectiveness of our extractor network.
With the proposed RL training procedure (BERT-ext + abs + RL),
our model exceeds the best model of \citet{chen-bansal-2018-fast}.
In addition, the result is better than those of all the other abstractive methods
exploiting extractive approaches in them
\cite{hsu-etal-2018-unified, chen-bansal-2018-fast, gehrmann-etal-2018-bottom}.

\paragraph{Redundancy Control}
Although the proposed RL training inherently gives training signals
that induce the model to avoid redundancy across sentences,
there can be still remaining overlaps between extracted sentences.
We found that the additional methods reducing redundancies can
improve the summarization quality, especially on CNN/Daily Mail dataset.

We tried Trigram Blocking \cite{liu2019fine} for extractor and
Reranking \cite{chen-bansal-2018-fast} for abstractor, and we
empirically found that the reranking only improves the performance.
This helps the model to compress the extracted sentences
focusing on disjoint information, even if there are some partial
overlaps between the sentences.
Our best abstractive model (BERT-ext + abs + RL + rerank) achieves
the new state-of-the-art performance for abstractive summarization
in terms of average ROUGE score, with large margins on ROUGE-L.

However, we empirically found that the reranking method
has no effect or has negative effect on NYT50 or DUC-2002 dataset.
Hence, we don't apply it for the remaining datasets.

\begin{table}
\begin{center}
\begin{adjustbox}{max width=0.95\columnwidth}
\begin{tabular}{l|ccc}
    \hline 
    \bf Models & \bf R-1 & \bf R-2 & \bf R-L \\ 
    \hline
    Sentence-level Reward & 40.82 & 18.63 & 38.41 \\
    Combinatorial Reward & 40.85 & 18.77 & 38.44 \\
    \hline
    Sentence-level Reward + rerank & 41.58 & 18.72 & 39.31 \\
    Combinatorial Reward + rerank & \bf 41.90 & \bf 19.08 & \bf 39.64 \\
    \hline
\end{tabular}
\end{adjustbox}
\end{center}
\caption{\label{table:ablation} Comparison of RL training.}
\end{table}

\paragraph{Combinatorial Reward}
Before seeing the effects of our summary-level rewards on final results,
we check the upper bounds of different training signals for the full model.
All the document sentences are paraphrased with our trained abstractor,
and then we find the best set for each search method.
\emph{Sentence-matching} finds sentences with the highest ROUGE-L score for
each sentence in the gold summary. This search method matches with the
best reward from \citet{chen-bansal-2018-fast}.
\emph{Greedy Search} is the same method explained for
extractor pre-training in section \ref{sec:submodule}.
\emph{Combination Search} selects a set of sentences
which has the highest summary-level ROUGE-L score,
from all the possible combinations of sentences.
Due to time constraints, we limited the maximum number
of sentences to 5. This method corresponds to our final return
in RL training.

Table \ref{table:oracle} shows the
summary-level ROUGE scores of previously explained methods.
We see considerable gaps between Sentence-matching and Greedy Search,
while the scores of Greedy Search are close to those of Combination Search.
Note that since we limited the number of sentences for Combination Search,
the exact scores for it would be higher.
The scores can be interpreted to be upper bounds for corresponding training methods.
This result supports our training strategy; pre-training with
Greedy Search and final optimization with the combinatorial return.

Additionally, we experiment to verify the contribution of our training method.
We train the same model with different training signals;
Sentence-level reward from \citet{chen-bansal-2018-fast} and combinatorial 
reward from ours. The results are shown in Table \ref{table:ablation}.
Both with and without reranking, the models trained with the combinatorial reward
consistently outperform those trained with the sentence-level reward.

\begin{table}
\begin{center}
\begin{adjustbox}{max width=\columnwidth}
\begin{tabular}{l|cc|c}
    \hline 
    \bf Models & \bf Relevance & \bf Readability & \bf Total \\ 
    \hline
    Sentence Rewrite \cite{chen-bansal-2018-fast} & 56 & 59 & 115 \\
    BERTSUM \cite{liu2019fine} & 58 & 60 & 118 \\
    BERT-ext + abs + RL + rerank (ours) & \bf 66 & \bf 61 & \bf 127 \\
    \hline
\end{tabular}
\end{adjustbox}
\end{center}
\caption{\label{table:human} Results of human evaluation.}
\end{table}

\paragraph{Human Evaluation}
We also conduct human evaluation to ensure robustness of our training procedure. We measure
relevance and readability of the summaries. Relevance is based on the summary containing important, salient information from the input article,
being correct by avoiding contradictory/unrelated
information, and avoiding repeated/redundant information. Readability is based on the summarys fluency, grammaticality, and coherence. To
evaluate both these criteria, we design a Amazon Mechanical Turk experiment based on ranking method, inspired by \citet{kiritchenko-mohammad-2017-best}.
We randomly select 20 samples from the CNN/Daily Mail test set and
ask the human testers (3 for each sample) to rank summaries (for relevance and readability) produced by 3 different models:
our final model, that of \citet{chen-bansal-2018-fast} and  that of \citet{liu2019fine}.
2, 1 and 0 points were given according to the ranking.
The models were anonymized and randomly shuffled.
Following previous work,
the input article and ground truth summaries are
also shown to the human participants in addition
to the three model summaries.
From the results shown in Table \ref{table:human}, we can see that our model is
better in relevance compared to others. In terms of readability, there was no
noticeable difference.

\begin{table}
\begin{center}
\begin{adjustbox}{max width=\columnwidth}
\begin{tabular}{l|ccc}
    \hline 
    \bf Models & \bf R-1 & \bf R-2 & \bf R-L \\ 
    \hline
    \multicolumn{4}{l}{Extractive} \\
    \hline
    First sentences \cite{durrett-etal-2016-learning} & 28.60 & 17.30 & - \\
    First $k$ words \cite{durrett-etal-2016-learning} & 35.70 & 21.60 & - \\
    Full \cite{durrett-etal-2016-learning} & 42.20 & 24.90 & - \\
    BERTSUM \cite{liu2019fine} & \bf 46.66 & 26.35 & 42.62 \\
    \hline
    \multicolumn{4}{l}{Abstractive} \\
    \hline
    Deep Reinforced \cite{paulus2018a} & 42.94 & 26.02 & - \\
    Two-Stage BERT \cite{zhang2019pretraining} & 45.33 & 26.53 & - \\
    BERT-ext + abs (ours) & 44.41 & 24.61 & 41.40 \\
    BERT-ext + abs + RL (ours) & \bf 46.63 & \bf 26.76 & \bf 43.38 \\
    \hline
\end{tabular}
\end{adjustbox}
\end{center}
\caption{\label{table:nyt50} Performance on NYT50 test set using the
limited length ROUGE recall score.}
\end{table}

\subsection{New York Times corpus}
Table \ref{table:nyt50} gives the results on NYT50 dataset.
We see our BERT-ext + abs + RL outperforms all the extractive
and abstractive models, except ROUGE-1 from \citet{liu2019fine}.
Comparing with two recent models that adapted BERT on their
summarization models \cite{liu2019fine, zhang2019pretraining},
we can say that we proposed another method successfully leveraging
BERT for summarization. In addition, the experiment proves
the effectiveness of our RL training, with about 2 point
improvement for each ROUGE metric.

\subsection{DUC-2002}
We also evaluated the models trained on the CNN/Daily Mail dataset
on the out-of-domain DUC-2002 test set as shown in Table \ref{table:duc2002}.
BERT-ext + abs + RL outperforms baseline models with large margins on
all of the ROUGE scores. This result shows that our model generalizes better.

\begin{table}
\begin{center}
\begin{adjustbox}{max width=\columnwidth}
\begin{tabular}{l|ccc}
    \hline 
    \bf Models & \bf R-1 & \bf R-2 & \bf R-L \\ 
    \hline
    Pointer Generator \cite{see-etal-2017-get} & 37.22 & 15.78 & 33.90 \\
    Sentence Rewrite \cite{chen-bansal-2018-fast} & 39.46 & 17.34 & 36.72 \\
    BERT-ext + abs + RL (ours) & \bf 43.39 & \bf 19.38 & \bf 40.14 \\
    \hline
\end{tabular}
\end{adjustbox}
\end{center}
\caption{\label{table:duc2002} Performance on DUC-2002 test set using
the full length ROUGE $F_1$ score.}
\end{table}

\section{Related Work}


There has been a variety of deep neural network models for
abstractive document summarization. One of the most dominant
structures is the sequence-to-sequence (seq2seq) models with attention mechanism 
\cite{rush-etal-2015-neural, chopra-etal-2016-abstractive, nallapati-etal-2016-abstractive}.
\citet{see-etal-2017-get} introduced Pointer Generator network
that implicitly combines the abstraction with the extraction,
using copy mechanism \cite{gu-etal-2016-incorporating, zeng2016efficient}.
More recently, there have been several studies that have attempted to improve
the performance of the abstractive summarization by explicitly combining them
with extractive models.
Some notable examples include the use of inconsistency loss \cite{hsu-etal-2018-unified},
key phrase extraction \cite{li-etal-2018-guiding, gehrmann-etal-2018-bottom}, and
sentence extraction with rewriting \cite{chen-bansal-2018-fast}.
Our model improves Sentence Rewriting with BERT as an extractor and summary-level rewards
to optimize the extractor.

Reinforcement learning has been shown to be effective to directly optimize
a non-differentiable objective in language generation including text summarization
\cite{DBLP:journals/corr/RanzatoCAZ15, DBLP:conf/iclr/BahdanauBXGLPCB17, paulus2018a, celikyilmaz-etal-2018-deep, narayan-etal-2018-ranking}.
\citet{DBLP:conf/iclr/BahdanauBXGLPCB17} use actor-critic methods
for language generation, using reward shaping \cite{Ng:1999:PIU:645528.657613}
to solve the sparsity of training signals. Inspired by this,
we generalize it to sentence extraction to give per step reward
preserving optimality.


\section{Conclusions}
We have improved Sentence Rewriting approaches for abstractive summarization,
proposing a novel extractor architecture exploiting BERT and a novel training
procedure which globally optimizes summary-level ROUGE metric.
Our approach achieves the new state-of-the-art on both CNN/Daily Mail and New York Times
datasets as well as much better generalization on DUC-2002 test set.

\section*{Acknowledgments}

We thank anonymous reviewers for their constructive and
fruitful comments.
This work was supported by the National
Research Foundation of Korea (NRF) grant funded by the
Korea government (MSIT) (NRF2016M3C4A7952587).


\clearpage
\appendix

\begin{table*}
\centering
\begin{adjustbox}{width=\textwidth}
\begin{tabular}{|p{15cm}|}
    \hline 
    \bf Source document \\
    \hline
(CNN) \textcolor{blue}{Duke University students and faculty members marched Wednesday afternoon chanting ``We are not afraid. We stand together,"} \textcolor{red}{after a noose was found hanging from a tree on campus.}
\textcolor{cyan}{Duke officials have asked anyone with information about the rope noose, which was found near a student center at 2 a.m., to call campus police.}
Photos of the noose prompted outrage from the community as they were passed along on social media.
At a forum held on the steps of Duke Chapel, close to where the noose was hung, hundreds of people gathered.
``You came here for the reason that you want to say with me, `This is no Duke we will accept. This is no Duke we want. This is not the Duke we're here to experience. And this is not the Duke we're here to create,' " Duke President Richard Brodhead told the crowd.
\textcolor{orange}{The incident is one of several recent racist events to affect college students.}
Last month a fraternity at the University of Oklahoma had its charter removed after a video surfaced showing members using the N-word and referring to lynching in a chant. Two students were expelled.
In February, a noose was hung around the neck of a statue of a famous civil rights figure at the University of Mississippi.
A statement issued by Duke said there was a previous report of hate speech directed at students on campus.
In the news release, the vice president for student affairs called the noose incident a ``cowardly act."
``To whomever committed this hateful and stupid act, I just want to say that if your intent was to create fear, it will have the opposite effect," Larry Moneta said.
Duke University is a private college with about 15,000 students in Durham, North Carolina. \\
   \hline
   \bf Ground truth summary \\
   \hline
   The noose, made of rope, was discovered on campus about 2 a.m.\\
   Hundreds of people gathered Wednesday afternoon to show solidarity against racism\\
   Duke official says to unknown perpetrator: You wanted to create fear but the opposite will happen\\
   \hline
   \bf BERT-ext + abs + RL + rerank \\
   \hline
   \textcolor{red}{A noose was found hanging from a tree on campus.}\\
   \textcolor{cyan}{Duke officials have asked anyone with information about the rope noose.}\\
   \textcolor{orange}{The incident is one of several racist events to affect college students.}\\
   \textcolor{blue}{Duke University students and faculty members marched Wednesday.}\\
   \hline
\end{tabular}
\end{adjustbox}
\caption{\label{table:example1} Example from the CNN/Dail Mail test set showing the generated summary of our best model. The colored sentences in the source document are the corresponding extracted sentences.}
\end{table*}

\clearpage

\begin{table*}
\centering
\begin{adjustbox}{width=\textwidth}
\begin{tabular}{|p{15cm}|}
    \hline 
    \bf Source document \\
    \hline
(CNN) \textcolor{red}{A SkyWest Airlines flight made an emergency landing in Buffalo, New York, on Wednesday after a passenger lost consciousness, officials said.}
\textcolor{blue}{The passenger received medical attention before being released, according to Marissa Snow, spokeswoman for SkyWest.} \textcolor{green}{She said the airliner expects to accommodate the 75 passengers on another aircraft to their original destination -- Hartford, Connecticut -- later Wednesday afternoon.}
\textcolor{cyan}{The Federal Aviation Administration initially reported a pressurization problem and said it would investigate.} Snow said there was no indication of any pressurization issues, and the FAA later issued a statement that did not reference a pressurization problem.
SkyWest also said there was no problem with the plane's door, which some media initially reported.
Flight 5622 was originally scheduled to fly from Chicago to Hartford. \textcolor{orange}{The plane descended 28,000 feet in three minutes.}
``It would feel like a roller coaster -- when you're coming over the top and you're going down," CNN aviation analyst Mary Schiavo said, describing how such a descent would feel. ``You know that these pilots knew they were in a very grave and very serious situation." \\
   \hline
   \bf Ground truth summary \\
   \hline
   FAA backtracks on saying crew reported a pressurization problem.\\
   One passenger lost consciousness.\\
   The plane descended 28,000 feet in three minutes.\\
   \hline
   \bf BERT-ext + abs + RL + rerank \\
   \hline
   \textcolor{red}{A SkyWest Airlines flight made an emergency landing in New York.}\\
   \textcolor{cyan}{FAA initially reported a pressurization problem.}\\
   \textcolor{orange}{The plane descended 28,000 feet in three minutes.}\\
   \textcolor{blue}{The passenger received medical attention before being released.}\\
   \textcolor{green}{The airliner expects to accommodate 75 passengers on another aircraft.}\\
   \hline
\end{tabular}
\end{adjustbox}
\caption{\label{table:example2} Example from the CNN/Dail Mail test set showing the generated summary of our best model. The colored sentences in the source document are the corresponding extracted sentences.}
\end{table*}

\end{document}